\documentclass{article}

\usepackage[preprint]{iaseai26}

\usepackage[utf8]{inputenc} %
\usepackage[T1]{fontenc}    %
\usepackage{hyperref}       %
\usepackage{url}            %
\usepackage{booktabs}       %
\usepackage{amsfonts}       %
\usepackage{nicefrac}       %
\usepackage{microtype}      %
\usepackage{xcolor}         %
\usepackage{bbm}
\usepackage{tcolorbox}
\usepackage{subcaption}

\usepackage{amsmath}
\usepackage[capitalize,noabbrev]{cleveref}
\usepackage{amssymb}
\usepackage{mathtools}
\usepackage{amsthm}

\newcommand{\set}[1]{\left\{ #1 \right\}}

\newcommand{\gpticon}{\includegraphics[height=1em]{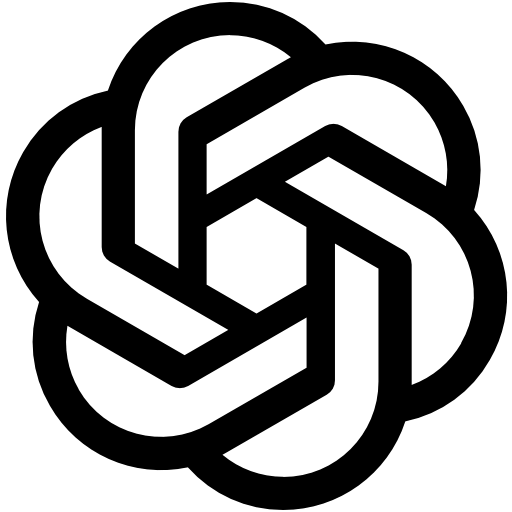}}
\newcommand{\llamaicon}{\includegraphics[height=1em]{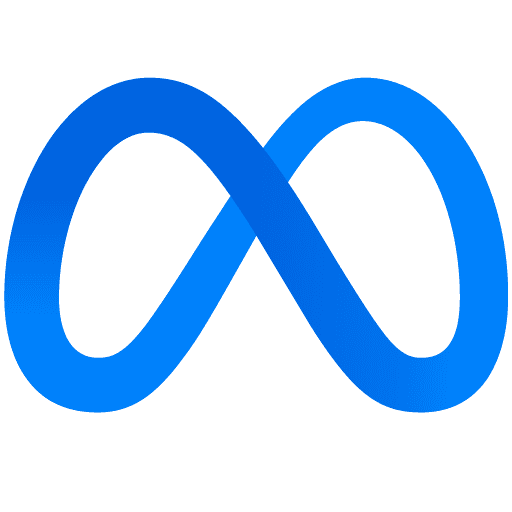}}
\newcommand{\mistralicon}{\includegraphics[height=1em]{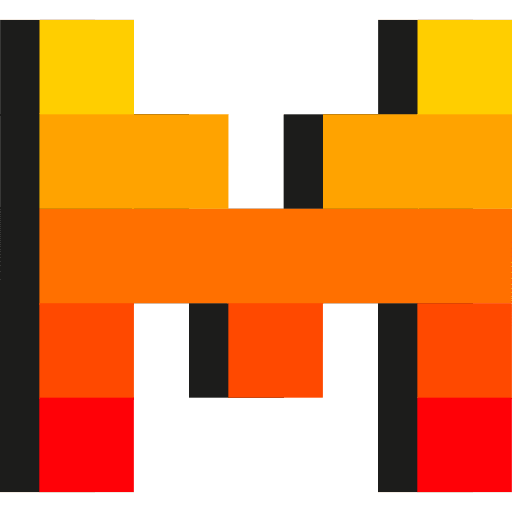}}

\definecolor{DarkGreen}{RGB}{85,146,38}

\title{Talk Isn’t Always Cheap: \\ Understanding Failure Modes in Multi-Agent Debate}

\author{%
  Andrea Wynn\thanks{equal contribution} \\
  Department of Computer Science\\
  Johns Hopkins University\\
  \texttt{awynn13@jhu.edu}\\
  \And
  Harsh Satija\footnotemark[1] \\
  Vector Institute \\
  \And 
  Gillian K. Hadfield \\
  Department of Computer Science\\
  Johns Hopkins University\\
}

\begin{document}

\maketitle

\begin{abstract}
While multi-agent debate has been proposed as a promising strategy for improving AI reasoning ability, we find that debate can sometimes be harmful rather than helpful.
Prior work has primarily focused on debates within homogeneous groups of agents, whereas we explore how diversity in model capabilities influences the dynamics and outcomes of multi-agent interactions. Through a series of experiments, we demonstrate that debate can lead to a decrease in accuracy over time — even in settings where stronger (i.e., more capable) models outnumber their weaker counterparts.
Our analysis reveals that models frequently shift from correct to incorrect answers in response to peer reasoning, favoring agreement over challenging flawed reasoning. We perform additional experiments investigating various potential contributing factors to these harmful shifts -- including sycophancy, social conformity, and model and task type. 
These results highlight important failure modes in the exchange of reasons during multi-agent debate, suggesting that naive applications of debate may cause performance degradation when agents are neither incentivised nor adequately equipped to resist persuasive but incorrect reasoning.
\end{abstract}

\section{Introduction}
\label{sec:intro}

Large Language Model (LLM) agents have demonstrated remarkable problem-solving abilities across a wide array of complex reasoning tasks \citep{brown2020language}.
Recently, a new line of research on interactive reasoning among multiple LLMs through debate has promoted the multi-agent debate framework as a promising approach to enhancing the reasoning and decision-making capabilities of LLM agents \citep{du2023improvingfactualityreasoninglanguage, chan2023chateval, liang2023encouraging, khan2024debatingpersuasivellmsleads}. Various forms of multi-agent debate have been shown to improve performance on multiple arithmetic and strategic reasoning benchmarks \citep{du2023improvingfactualityreasoninglanguage, subramaniam2025multiagent}, produce more truthful answers and evaluations \citep{chan2023chateval, khan2024debatingpersuasivellmsleads}, and enhance tasks such as machine translation \citep{liang2023encouraging} and negotiation \citep{fu2023improving}.
The core concept of these studies is that by engaging LLM agents through structured argumentation or discourse, we can facilitate the exchange of reasoning among different agents and guide them toward more accurate answers. Intuitively, greater exchanges of reasoning should lead to better decisions—allowing multiple agents to challenge flawed reasoning, highlight overlooked details, and reduce individual biases. But is this always the case?

In this work, we show that the benefits of multi-agent debate are not as universal as commonly assumed. Through a series of empirical studies, we show that multi-agent debate can sometimes degrade performance, leading to worse final answers than those generated by a single agent acting alone. These failures are not rare edge cases, but arise systematically in settings where agents amplify each other’s errors -- agreeing reflexively rather than challenging flawed reasoning.
These findings hold even when there is variation in the abilities of the participating LLM agents. For instance, we discover that introducing a weak or less capable (lower-performing) LLM agent into a debate with a strong or more capable (higher-performing) agent can detrimentally affect the debate outcome, producing results worse than if the agents had not engaged in discussion. The presence of a weaker agent disrupts the performance of the stronger agent. Moreover, in certain cases, the longer a debate continues, the more performance can degrade.
In other words: talk isn’t always cheap -- and in some cases, it’s actively harmful.

We present a systematic evaluation of multi-agent debate across multiple tasks, showing that debate can sometimes \textit{harm} group performance, particularly with heterogeneous LLM agents engaged in debate.
Our findings challenge the prevailing narrative that more discussion between agents is inherently beneficial. Instead, we uncover several key factors that mediate the success or failure of debate, including task type and complexity, agent diversity and capability, and social influence. In doing so, we offer a nuanced view of when and why debate helps -- and when it hurts.
Together, these results suggest that while debate remains a promising tool for improving model reasoning, it should be applied carefully to ensure safety on the task and setting of interest. 

\textbf{Statement of Contributions.} We summarize our contributions as follows:
\begin{enumerate}
    \item We conduct a comprehensive evaluation of the effectiveness of the multi-agent debate framework across three different datasets. In \Cref{sec:effectiveness-of-debate}, we demonstrate that debate may degrade performance compared to majority voting. Additionally, in \Cref{sec:performance-degradation-during-debate} we show that the performance may progressively deteriorate as the debate progresses.
    \item We extend the debate framework beyond the homogeneous setting to investigate the effect of diverse agent populations.  Contrary to the prevalent belief that LLM agents are inherently collaborative -- suggesting that a mixture of diverse models improves response quality when they can access outputs from other models, even if those outputs are of lower quality \citep{wang2024mixture} -- we find this assumption does not hold in multi-agent debates. In \Cref{sec:results}, we observe that the presence of weaker agents can negatively affect performance.
    \item Our analysis in \Cref{sec:failure-modes-of-debate} reveals that a significant portion of correct answers become corrupted during debate. We investigate this phenomenon from various perspectives, such as the effect of sequential revision (\Cref{sec:sequential-revision}), social influence (\Cref{sec:social-factors}), and sycophancy (\Cref{sec:investigation-debate-harm-factors}). These insights opens avenues for future research focused on enhancing reasoning exchange in multi-agent systems.
\end{enumerate}

\section{Related Work}

\textbf{Debate and multi-agent reasoning.}
Multi-agent debate was initially proposed as method for the scalable oversight problem  where a judge or verifier can interject and elicit hidden contradictions, using structured back-and-forth conversations \citep{irving2018aisafetydebate, khan2024debatingpersuasivellmsleads, michael2023debatehelpssuperviseunreliable, kenton2024scalableoversightweakllms}. More recently, another form of multi-agent debate, sometimes referred to as multi-agent deliberation, investigates leveraging different LLM agents to surface better answers by having them exchange reasoning via iterative discussion \citep{chan2023chateval, liang2023encouraging, subramaniam2025multiagent}.  Most of these studies focus on a homogeneous setting where all LLM agents utilize the same underlying language model \citep{du2023improvingfactualityreasoninglanguage} or a model of similar ability \citep{yao2025peacemakertroublemakersycophancyshapes}, finding that this approach enhances accuracy across various Question-Answering (QA) tasks.

\citet{estornell2024multi} examine the theoretical properties and effects of opinion diversity within the debate framework, reporting a ``tyranny of the majority'' effect. They found that if the majority of agents provide the same answer -- regardless of its correctness -- minority agents tend to conform, creating an echo chamber effect. Additionally, they propose a theoretical result indicating that diversity of model abilities should \textit{improve} overall debate performance; we show empirical evidence that this is often not true in practice. 

Some works explore debates between agents of diverse nature or abilities. For instance,  \citet{estornell2025acccollabactorcriticapproachmultiagent, subramaniam2025multiagent} propose training LLM agents to debate collaboratively with distinct roles (actors/generators and critics), demonstrating that this approach can surpass previous unsupervised debate setups in reasoning benchmarks.
Finally, studies such as \citet{amayuelas2024multiagent} investigate whether the collaborative nature holds if an explicit adversary is introduced into the debate process—where the adversary actively seeks to reduce performance. These works suggest that, when agents constructively challenge each other, answers can improve.
However, other studies caution that debate can fail when agents emphasize persuasion over truth. \citet{agarwal2025persuasionoverridestruthmultiagent} introduce a single-round debate on factual questions (using TruthfulQA) where one agent states a true answer calmly and another delivers a confident, emotional false answer. They show the LLM judge often chooses the persuasive falsehood with high confidence, suggesting that a vivid but incorrect argument can override a correct one. These results highlight a risk: unless the judge is well-calibrated, debate may amplify bluster, mirroring human misinformation scenarios.
\citet{yao2025peacemakertroublemakersycophancyshapes} focus on investigating sycophancy in the debate setting, showing that agent disagreement rate decreases as debate progresses, and that this observation is correlated with performance degradation. 

\textbf{Collaborative multi-agent frameworks.} Beyond explicit debates, many recent systems assume collaboration among LLMs improves reasoning \citep{li2023camelcommunicativeagentsmind, wang2024mixture, tran2025multi}. For instance, \citet{wu2023autogenenablingnextgenllm} provides a general multi-agent conversation framework: developers can define many agents (assistant, user-proxy, tools, etc.) that autonomously chat to solve complex tasks. These role-based and decentralized systems often yield richer interactions (e.g. multi-turn planning) than a single LLM alone.

\textbf{Sequential revision.}
Frameworks that utilize interactive reasoning at inference time in a sequential manner are employed for either self-refinement \citep{madaan2023self} or self-consistency \citep{wang2022self}. Self-refinement involves iteratively revising or adapting a model's responses based on previous outputs, prompting the model to intentionally reflect on its existing responses and correct any mistakes \citep{kamoi2024can}.
Works based on self-consistency often explicitly run multiple reasoning paths in parallel. For example, \citet{wang2023selfconsistencyimproveschainthought} sample numerous independent chain-of-thought answers and select the most common answer. This aggregation of diverse reasoning paths significantly enhances the accuracy of arithmetic and commonsense QA by minimizing uncertainty.
\citet{he2025enhancingllmreasoningmultipath} extend this idea: they spawn multiple ``reactive'' and ``reflection'' agent pairs, each exploring a different reasoning path, and then use a separate summarizer to aggregate them. Likewise, \citet{yang2025llmpowereddecentralizedgenerativeagents} build a decentralized multi-agent planner where each LLM agent maintains its own memory (a hierarchical knowledge graph) and communicates via structured prompts. They find that these collaborative agents reach goals with ~60\% fewer steps than a lone agent, underscoring the value of structured cooperation.

Collectively, prior work shows that multi-agent debate and collaboration can amplify reasoning abilities, but also highlights key failure modes: judges may be fooled by rhetoric, and human-like dynamics can bias group outputs. Our work builds on these insights by expanding the analysis to heterogeneous debate settings, then performing a deep dive into the potential factors affecting when debate helps -- or hurts -- LLM performance on a variety of tasks.

\section{Setting: Multi-agent debate}
\label{sec:setting}

Let $\mathcal{Q}$ denote a dataset of questions related to a task, where each question $q \in \mathcal{Q}$ in the task is natural language text. The objective is to generate an answer $a \in \mathcal{A}$ for any given input question $q \in \mathcal{Q}$, where $\mathcal{A}$ is the set of possible answers for that task.
We assume that there exists a ground truth answer $a^\star \in \mathcal{A}$ for each question $q \in \mathcal{Q}$ which is denoted by $f^{gt}: \mathcal{Q} \rightarrow \mathcal{A}$, i.e., $f^{gt}(q) = a^\star$.

\noindent{\textbf{Single-Agent Setting}}:
In the single agent setting, each agent uses an underlying LLM $l : \mathcal{Q} \rightarrow \mathcal{A}$ to generate answer $a \sim l(q)$ where $a \in \mathcal{A}$ denotes the answer generated by the LLM in response to the question $q$. The main metric of interest is accuracy which is calculated as $\frac{1}{|\mathcal{Q}|} \sum_{q \in \mathcal{Q}} \mathbbm{1}[l(q) = f^{gt}(q)]$.

Usually, various task specific prompts are also given as input to the LLM in addition to the input question for generating an answer. Let $\mathcal{P}_{\text{T}}$ denote the task-specific prompt, then the resulting LLM can be described as $l_{\mathcal{P}_{\text{T}}} : \mathcal{P}_{\text{T}} \times \mathcal{Q} \rightarrow \mathcal{A}$ or $a \sim l(\mathcal{P}_{\text{T}}(q))$.

\noindent{\textbf{Multi-Agent Debate}}:
We follow the multi-agent debate framework from \cite{du2023improvingfactualityreasoninglanguage,subramaniam2025multiagent} which involves initially posing a question to a group of LLM agents. After the initial responses, which include answers plus reasoning, are generated, the debate process then iteratively revisits the question for each agent, contextualizing input through the collective responses from previous rounds. Specifically, each agent generates a new response to the question based on its prior response and the summarized responses from the other agents. The final answer is based on the majority vote among all agents.

Formally, we have a group of $N$ agents, each with their own LLM $l_i$, that are presented with a question $q \in \mathcal{Q}$ and tasked with generating an answer $a \in \mathcal{A}$. We use $d(l_1, \ldots, l_n): \mathcal{Q} \rightarrow \mathcal{A}$ to denote the debate procedure that takes as input a question and a set of LLM agents and generates an answer. The debate procedure runs over multiple rounds, where in any given round $t$, each agent $i$ uses their underlying LLM $l_i$ to iteratively generate an answer in the following manner:
\begin{enumerate}
    \item \textbf{Starting Round ($t=0$):} Each agent $i$ generates a response $g_i \in \mathcal{G}$ via $l_i$ using a starting prompt $\mathcal{P}_{\texttt{starting}}$, i.e.
    $g_i^0 \sim l_i(\mathcal{P}_{\texttt{starting}}(q))$. Here $\mathcal{G}$ denotes the set of possible generations from LLM.

    \item \textbf{Debate Rounds ($t=1, \ldots, T$):} For the subsequent rounds, each agent $i$ is given the question $q$ and responses from the other agents from the previous round.
          Let $o_i^{t} = \set{ g_j^{t-1} }_{j \neq i }$ denote the outputs from the other agents from previous round. Then the goal of agent $i$ for this round is to generate an updated response $g^t_i \in \mathcal{G}$ that takes into account the responses from the other agents and its own response from the previous round $g^{t-1}_i$, i.e.,
          \begin{equation*}
              g^t_i \sim l_i(\mathcal{P}_{\text{debate}}(q, o^t_i, g^{t-1}_i)),
          \end{equation*}
          where $\mathcal{P}_{\text{debate}}$ is the debate prompt.~\footnote{If the responses are too large to fit in the context window, then they are summarized via making another LLM call, i.e., $o_i^{t} = \set{ l_i(\mathcal{P}_{\text{summarize}}(g_j^{t-1}))}_{j \neq i }$.}
\end{enumerate}
This procedure runs for $T$ rounds, where the final output of the debate is the set of responses $\{g^T_i\}_{i=1}^n$. Then the majority response across all agents is selected as the final answer. Let $\texttt{majority} : \mathcal{G}^N \rightarrow \mathcal{A}$ denote the majority voting and filtering for responses across all agents, then the final answer is given by $a = \texttt{majority}(\{g^T_i\}_{i=1}^n)$. Accuracy is evaluated with respect to the final for the group of debating agents.

\section{Experimental Setup}

In this section, we describe the set of tasks, models, and prompts used in our experiments.

\subsection{Datasets}

\textbf{CommonSenseQA (CSQA):} The CommonSenseQA dataset \citep{commonsenseqa} consists of multiple-choice questions with complex semantics that often require prior knowledge to answer correctly. The dataset is intended to test for prior common-sense knowledge encoded within LLMs and checks for common misconceptions. Additionally, this task (common-sense reasoning) was not evaluated in prior work on multi-agent debate, making it a good testbed for evaluating generalization of these debate approaches to different tasks. 

\textbf{MMLU:} Massive Multitask Language Understanding, or MMLU \citep{mmlu}, is a widely used multiple-choice dataset covering 57 domains including elementary mathematics, US history, computer science, law, and more. To perform well on MMLU, models need robust world knowledge and problem solving ability.

\textbf{GSM8K:} GSM8K \citep{gsm8k} is a dataset of linguistically diverse grade school math word problems which require multi-step mathematical reasoning to solve. This dataset is not multiple-choice and instead requires open-ended generation of the answer to the math questions, potentially with intermediate reasoning steps.

\subsection{Models}

We used three models from distinct model families in our experiments: GPT-4o-mini \citep{openai2024gpt4omini}, LLaMA-3.1-8B-Instruct \citep{llama3} and Mistral-7B-Instruct-v0.2 \citep{mistral}. To align with prior work on multi-agent debate \citep{du2023improvingfactualityreasoninglanguage, subramaniam2025multiagent}, we ran all experiments and models with the default temperature parameter, the $\texttt{top\_p} = 0.9$, maximum generation length of $2048$ tokens, and $T=2$ rounds of debate.
We take 100 random samples for each task from the dataset and report the result over 5 random seeds.

\subsection{Prompts}

We provide prompts on an example question for each task below: 

\begin{itemize}
    \item \textbf{CommonSenseQA:} Can you answer the following question as accurately as possible? If a product doesn't last, what does it have a reputation of doing?: A) disintegrating, B) wearing out, C) dissolving, D) falling apart, E) dissipating Explain your answer by providing a bullet point summary of your reasoning, putting the answer in the form (X) at the end of your response.
    \item \textbf{MMLU:} Can you answer the following question as accurately as possible? What is the value of p in 24 = 2p?: A) p = 4, B) p = 8, C) p = 12, D) p = 24 Explain your answer by providing a bullet point summary of your reasoning, putting the answer in the form (X) at the end of your response.
    \item \textbf{GSM8K:} Can you solve the following math problem? Mark is trying to choose between two venues for a surprise party for his wife. The first venue charges a flat fee of \$200, regardless of how many guests attend; the second charges \$25 per person who attends.  However, the first venue does not include food, which Mark estimates will cost \$5 for each person who attends.  At the second venue, food for each guest is already included in the price.  How many guests are necessary for the two venues to be equal in cost? Provide a bullet point summary of your reasoning. Your final answer should be a single numerical number, in the form \boxed{answer}, at the end of your response.
\end{itemize}

For all models and tasks, we provide the following system prompt, $\mathcal{P}_{\text{system}}$:

\begin{tcolorbox}[
    colback=gray!5,
    colframe=gray!50,
    title={\textbf{System prompt}, $\mathcal{P}_{\text{system}}$},
    fonttitle=\bfseries,
    boxrule=1pt,
    arc=3pt,
    left=8pt,
    right=8pt,
    top=8pt,
    bottom=8pt
    ]
    You are a helpful assistant that can answer questions and provide helpful information.
\end{tcolorbox}

For multi-agent debate, we additionally use the following prompt as $\mathcal{P}_{\text{debate}}$ for each round of debate, adjusted to the answer format of each task (following \cite{du2023improvingfactualityreasoninglanguage}):

\begin{tcolorbox}[
    colback=gray!5,
    colframe=gray!50,
    title={\textbf{Debate prompt}, $\mathcal{P}_{\text{debate}}$},
    fonttitle=\bfseries,
    boxrule=1pt,
    arc=3pt,
    left=8pt,
    right=8pt,
    top=8pt,
    bottom=8pt
    ]
    These are the solutions to the problem from other agents: \{AGENT\_RESPONSES\} Using the reasoning from other agents as additional advice, can you give an updated answer? Explain your reasoning. Examine your solution and that of other agents. Put your answer in the form (X) at the end of your response.
\end{tcolorbox}

\subsection{Code}
We provide the source code for all our experiments at \url{https://github.com/TheNormativityLab/talk-aint-cheap/}.

\section{Results}
\label{sec:results}

\begin{table*}[t]
    \centering
    \resizebox{\textwidth}{!}{%
        \begin{tabular}{|l|c|c|c|c|c|c|}
            \hline
            \hline
                                                                             & \multicolumn{2}{c|}{\textbf{CommonSense QA}}    & \multicolumn{2}{c|}{\textbf{MMLU}}                       & \multicolumn{2}{c|}{\textbf{GSM8K}}                                                                                                                                                                 \\
            1$\times$ \gpticon                                               & \multicolumn{2}{c|}{74.8$\pm${\scriptsize 1.9}} & \multicolumn{2}{c|}{82.6$\pm${\scriptsize 2.1}}          & \multicolumn{2}{c|}{93.2$\pm${\scriptsize 1.8}}                                                                                                                                                     \\
            1$\times$ \llamaicon                                             & \multicolumn{2}{c|}{57.0$\pm${\scriptsize 1.5}} & \multicolumn{2}{c|}{55.6$\pm${\scriptsize 0.5}}          & \multicolumn{2}{c|}{76.4$\pm${\scriptsize 2.1}}                                                                                                                                                     \\
            1$\times$ \mistralicon                                           & \multicolumn{2}{c|}{41.6$\pm${\scriptsize 2.3}} & \multicolumn{2}{c|}{34.0$\pm${\scriptsize 1.9}}          & \multicolumn{2}{c|}{34.2$\pm${\scriptsize 1.8}}                                                                                                                                                     \\
            \hline
            \hline
                                                                             & w/o Debate                                      & After Debate                                             & w/o Debate                                      & After Debate                                              & w/o Debate                 & After Debate                                             \\
            \hline
            \hline
            3$\times$ \mistralicon                                           & 44.4$\pm${\scriptsize 2.7}                      & 39.4$\pm${\scriptsize 3.9} {\color{red}$\downarrow$ 5.0} & 33.6$\pm${\scriptsize 1.8}                      & 24.4$\pm${\scriptsize 2.9} {\color{red}$\downarrow$ 9.2}  & 43.6$\pm${\scriptsize 1.5} & 46.4$\pm${\scriptsize 1.4} {\color{green}$\uparrow$ 2.8} \\
            3$\times$ \llamaicon                                             & 63.0$\pm${\scriptsize 3.9}                      & 58.6$\pm${\scriptsize 2.3} {\color{red}$\downarrow$ 4.4} & 61.6$\pm${\scriptsize 2.4}                      & 57.8$\pm${\scriptsize 1.8} {\color{red}$\downarrow$ 3.8}  & 87.6$\pm${\scriptsize 1.5} & 84.2$\pm${\scriptsize 2.0} {\color{red}$\downarrow$ 3.4} \\
            3$\times$ \gpticon                                               & 75.6$\pm${\scriptsize 2.2}                      & 74.8$\pm${\scriptsize 2.1} {\color{red}$\downarrow$ 0.8} & 81.4$\pm${\scriptsize 3.3}                      & 82.2$\pm${\scriptsize 2.7} {\color{green}$\uparrow$ 0.8}  & 94.0$\pm${\scriptsize 0.9} & 94.4$\pm${\scriptsize 1.5} {\color{green}$\uparrow$ 0.4} \\
            \hline
            1$\times$ \gpticon, 2$\times$ \llamaicon                         & 66.2$\pm${\scriptsize 2.2}                      & 64.4$\pm${\scriptsize 2.1} {\color{red}$\downarrow$ 1.8} & 65.0$\pm${\scriptsize 2.3}                      & 68.0$\pm${\scriptsize 2.2} {\color{green}$\uparrow$ 3.0}  & 88.4$\pm${\scriptsize 1.3} & 92.8$\pm${\scriptsize 1.7} {\color{green}$\uparrow$ 4.4} \\
            2$\times$ \gpticon, 1$\times$ \llamaicon                         & 74.8$\pm${\scriptsize 1.3}                      & 74.0$\pm${\scriptsize 0.7} {\color{red}$\downarrow$ 0.8} & 82.6$\pm${\scriptsize 3.2}                      & 81.0$\pm${\scriptsize 3.0} {\color{red}$\downarrow$ 1.6}  & 93.6$\pm${\scriptsize 0.8} & 94.6$\pm${\scriptsize 1.3} {\color{green}$\uparrow$ 1.0} \\
            2$\times$ \llamaicon, 1$\times$ \mistralicon                     & 58.2$\pm${\scriptsize 3.8}                      & 50.2$\pm${\scriptsize 3.9} {\color{red}$\downarrow$ 8.0} & 51.8$\pm${\scriptsize 2.2}                      & 43.6$\pm${\scriptsize 1.9} {\color{red}$\downarrow$ 8.2}  & 82.6$\pm${\scriptsize 1.9} & 75.8$\pm${\scriptsize 2.1} {\color{red}$\downarrow$ 6.8} \\
            1$\times$ \llamaicon, 2$\times$ \mistralicon                     & 53.4$\pm${\scriptsize 2.7}                      & 46.8$\pm${\scriptsize 2.5} {\color{red}$\downarrow$ 6.6} & 40.0$\pm${\scriptsize 2.1}                      & 28.0$\pm${\scriptsize 1.2} {\color{red}$\downarrow$ 12.0} & 61.0$\pm${\scriptsize 2.4} & 64.8$\pm${\scriptsize 1.5} {\color{green}$\uparrow$ 3.8} \\
            1$\times$ \gpticon, 2$\times$ \mistralicon                       & 62.4$\pm${\scriptsize 1.1}                      & 59.4$\pm${\scriptsize 1.9} {\color{red}$\downarrow$ 3.0} & 65.8$\pm${\scriptsize 2.9}                      & 58.8$\pm${\scriptsize 1.2} {\color{red}$\downarrow$ 7.0}  & 90.2$\pm${\scriptsize 0.8} & 87.8$\pm${\scriptsize 1.9} {\color{red}$\downarrow$ 2.4} \\
            2$\times$ \gpticon, 1$\times$ \mistralicon                       & 74.6$\pm${\scriptsize 1.6}                      & 72.4$\pm${\scriptsize 2.7} {\color{red}$\downarrow$ 2.2} & 82.8$\pm${\scriptsize 2.7}                      & 80.8$\pm${\scriptsize 2.8} {\color{red}$\downarrow$ 2.0}  & 93.4$\pm${\scriptsize 1.3} & 93.0$\pm${\scriptsize 1.3} {\color{red}$\downarrow$ 0.4} \\
            1$\times$ \gpticon, 1$\times$ \llamaicon, 1$\times$ \mistralicon & 66.6$\pm${\scriptsize 1.9}                      & 65.4$\pm${\scriptsize 2.1} {\color{red}$\downarrow$ 1.2} & 57.8$\pm${\scriptsize 3.1}                      & 63.4$\pm${\scriptsize 2.1} {\color{green}$\uparrow$ 5.6}  & 86.8$\pm${\scriptsize 0.9} & 90.2$\pm${\scriptsize 1.2} {\color{green}$\uparrow$ 3.4} \\
            \hline
            \hline
        \end{tabular}
    }
    \caption{
        The first column shows the configuration of LLM agents, where \gpticon\,denotes GPT-4o-mini, \mistralicon\,denotes Mistral-7B and \llamaicon\,denotes the Llama-3.1-8B models.
        The top 3 rows benchmark the performance of how well a single agent performs on these tasks. The subsequent rows benchmark the effectiveness of debate procedure. The \textbf{w/o Debate} column represents the case where majority vote from multiple agents based on their initial responses is chosen, i.e., there is no exchange of reasoning.  The \textbf{After Debate} denotes the result of majority vote after the exchange of reasons via the debate procedure.
        The arrows denote the difference in performance post debate procedure characterizing the benefit of exchange of reasons on the performance. A \textcolor{red}{red arrow $\downarrow$} indicates a \textit{decrease} in performance after debate, while a \textcolor{green}{green arrow $\uparrow$} indicates an \textit{increase} in performance after debate. 
        All the experiments were done on 100 random samples and across 5 different seeds, reported are mean and standard error.
    }
    \label{tab:debate-effectiveness}
\end{table*}

\subsection{Effectiveness of Debate}
\label{sec:effectiveness-of-debate}

We present results showing that debate can sometimes be \textit{harmful} rather than helpful -- in particular, that sometimes agents perform better \textit{without any debate} than after exchanging reasons with other agents.
We present the results in \Cref{tab:debate-effectiveness}.
We find that in the case of CommonSenseQA, which was not studied in prior work on multi-agent debate \citep{du2023improvingfactualityreasoninglanguage}, debate always harms performance.
A key insight is that even when groups include more ``strong'' models (e.g., GPT) than ``weak'' ones (e.g., Mistral), the process of debate does not always yield performance gains.
Contrary to the prevailing narrative that debate improves collective reasoning, our experiments demonstrate that performance can actually decrease after agents engage in debate, even when stronger models outnumber weaker ones.

\subsection{Performance Degradation during Debate}
\label{sec:performance-degradation-during-debate}

\begin{figure*}[h!]
    \centering
    \includegraphics[width=\textwidth]{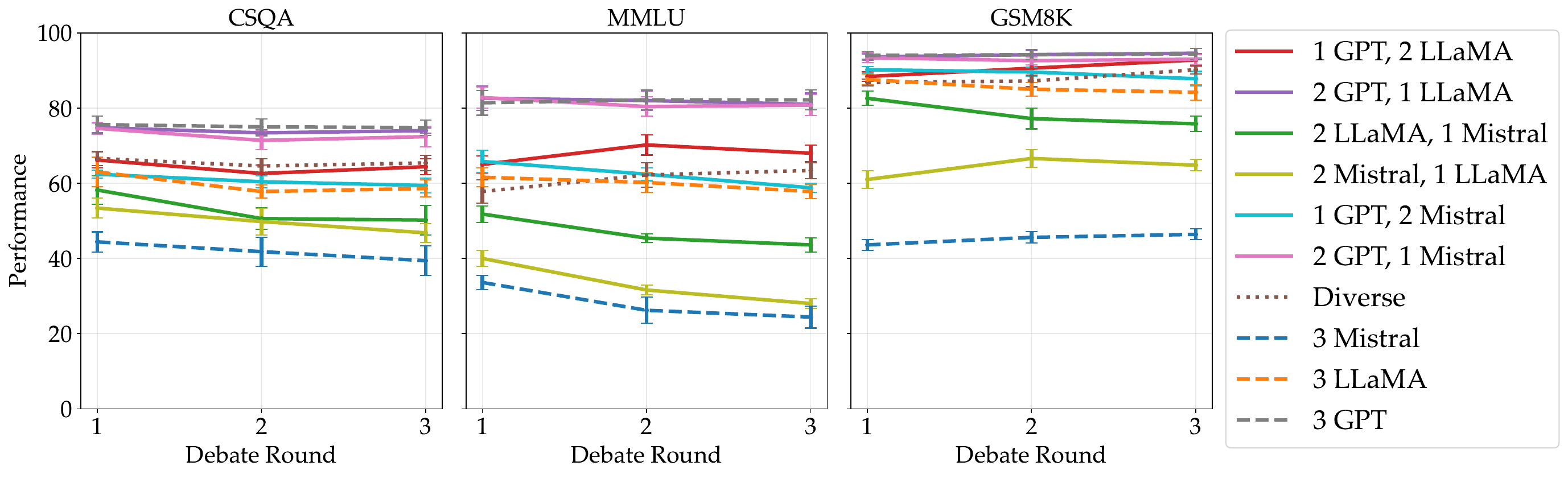}
    \caption{In many cases, we find that group accuracy frequently \textit{degrades} over the course of debate, rather than improving performance. Diverse refers to the case (1x\,\gpticon, 1x\,\llamaicon, 1x\,\mistralicon).
    }
    \label{fig:performance_vs_round}
\end{figure*}

\Cref{fig:performance_vs_round} presents performance across three tasks -- MMLU, CommonSenseQA, and GSM8K -- as a function of debate rounds among groups of language models with varying individual performance on the underlying task.
In fact, in many group configurations, we observe that performance \textit{decreases} as the debate progresses. This trend appears across datasets, but is especially pronounced in MMLU and CommonSenseQA, where groups with mixed-capability models often suffer from group performance degradation during debate despite having a majority of stronger agents.
These findings challenge the belief that deliberation or iterative reasoning among AI agents will always lead to better outcomes.

\section{Failure Modes of Debate}
\label{sec:failure-modes-of-debate}

Multi-agent debate is intended to improve reasoning by leveraging disagreement—encouraging models to refine their arguments through interaction and converge on correct conclusions. Yet, debates between LLM agents often fail to reach majority agreement on the true answer, even when at least one agent is initially correct. Examining why these failures occur is crucial for both understanding and designing these systems. We show that there are many other factors influencing failure modes of debate, which we explore in depth in this section. This suggests a richer and more complex interplay of factors influencing the effectiveness of debate. In this section, we dissect these dynamics empirically, examining when and how agents revise their answers, what social conditions are correlated with undesirable answer revisions, and whether an explicit intervention targeting sycophancy can meaningfully improve debate performance.

\subsection{Does exchange of reasoning help in sequential revision?}
\label{sec:sequential-revision}

If an LLM agent can reflect and correct mistakes based on the reasoning of other agents, we would expect the model to improve its answer and the collective performance of the group. We know the self-correction capability of single-agent LLMs does not easily work out of the box \citep{huang2023large}, and we want to evaluate if the debate procedure helps with this or not.

\begin{figure*}[h!]
    \centering
    \includegraphics[width=\textwidth]{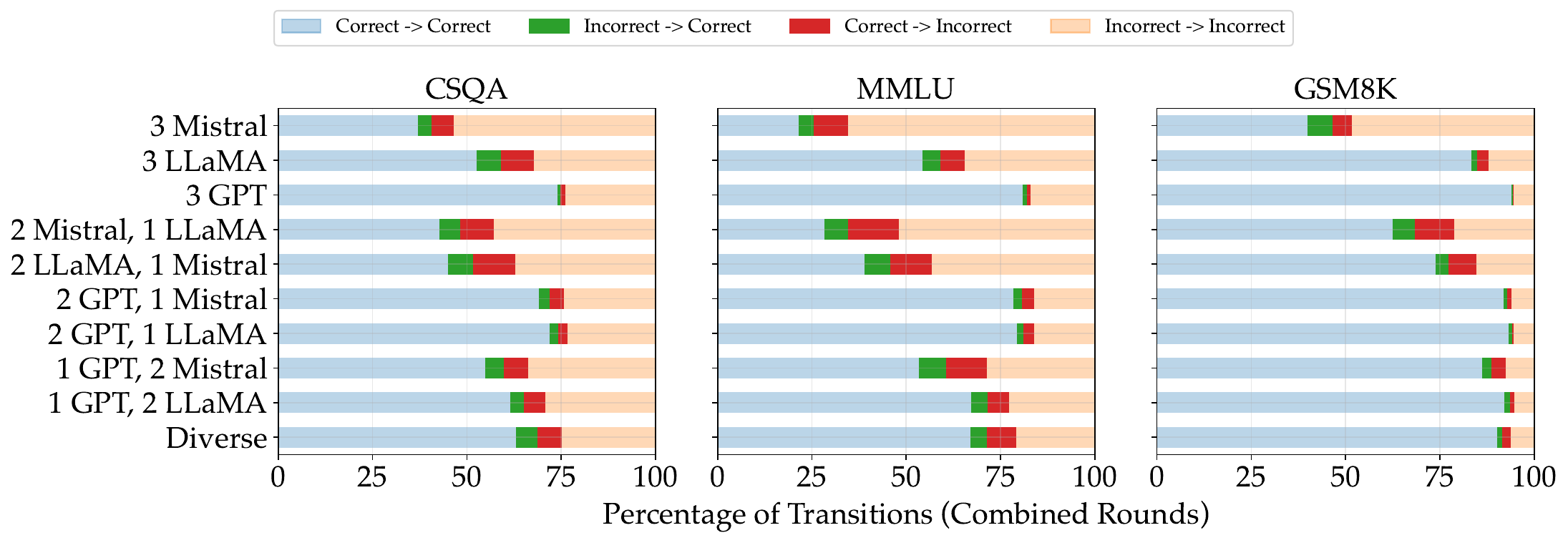}
    \caption{Breakdown of how agent change answers for different agent settings; results are aggregated over all debate rounds. We observe that most agents with incorrect initial answers do not improve their overall performance (peach bars), and, of those that do change their answers, more change from a correct answer to an incorrect one (red region) than from an incorrect to a correct one (green region). }
    \label{fig:combined-sycophancy}
\end{figure*}

\begin{figure*}[h!]
    \centering
    \includegraphics[width=\textwidth]{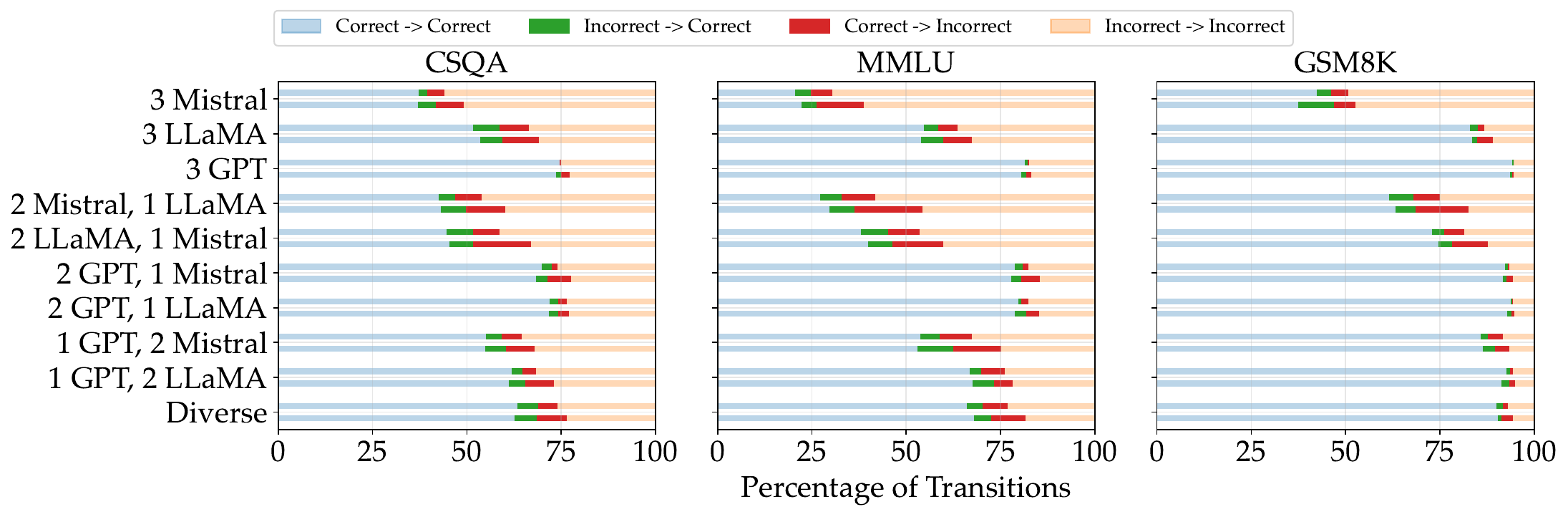}
    \caption{Breakdown of how agents change answers between debate rounds for different agent settings. The top row denotes the first round, and the row below denotes the second round. We find that the social effect dominates: agents that can resist flipping originally correct answers in round 1 have lower resistance to the social pressure from disagreement after round 2. }
    \label{fig:transitions-per-round}
\end{figure*}

To assess this, we analyze how agent responses change between debate rounds across all our tasks. Note that there are four possible types of transitions: correct $\rightarrow$ incorrect, correct $\rightarrow$ correct, incorrect $\rightarrow$ incorrect, and incorrect $\rightarrow$ correct. As shown in \Cref{fig:combined-sycophancy}, we observe that most agents with initially incorrect answers do not improve their performance (peach bars), and, of those that do change their answers, there exists a larger shift in agent responses from correct $\rightarrow$ incorrect answers (red) than incorrect $\rightarrow$ correct (green), indicating that debate can actively mislead agents who started with correct answers. 
Further, \Cref{fig:transitions-per-round} shows that the proportion of correct-to-incorrect transitions exceeds the proportion of incorrect-to-correct ones in subsequent rounds, corroborating our earlier results that debate performance degrades over rounds. We further find that there is a dominating effect of social pressure: agents that can originally resist flipping their correct answers to incorrect ones in round 1 have lower resistance to the social pressure from disagreement after round 2 (larger red bars in round 2). Together, these results highlight an important observation: to mitigate the harmful effects of debate, we need to find a way to reduce the number of \textit{undesirable} answer flips -- i.e., models changing their answers from correct $\rightarrow$ incorrect -- and address the underlying social effects that cause these flips. 

\subsection{Are agents influenced by social factors?}
\label{sec:social-factors}

\begin{figure}
    \centering
    \includegraphics[height=0.03\textheight]{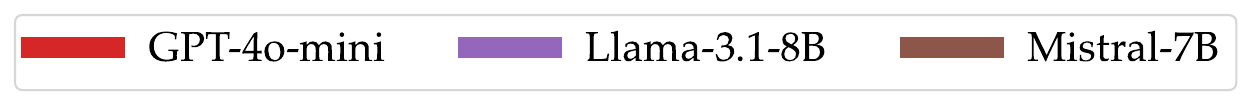}
    \includegraphics[width=\textwidth]{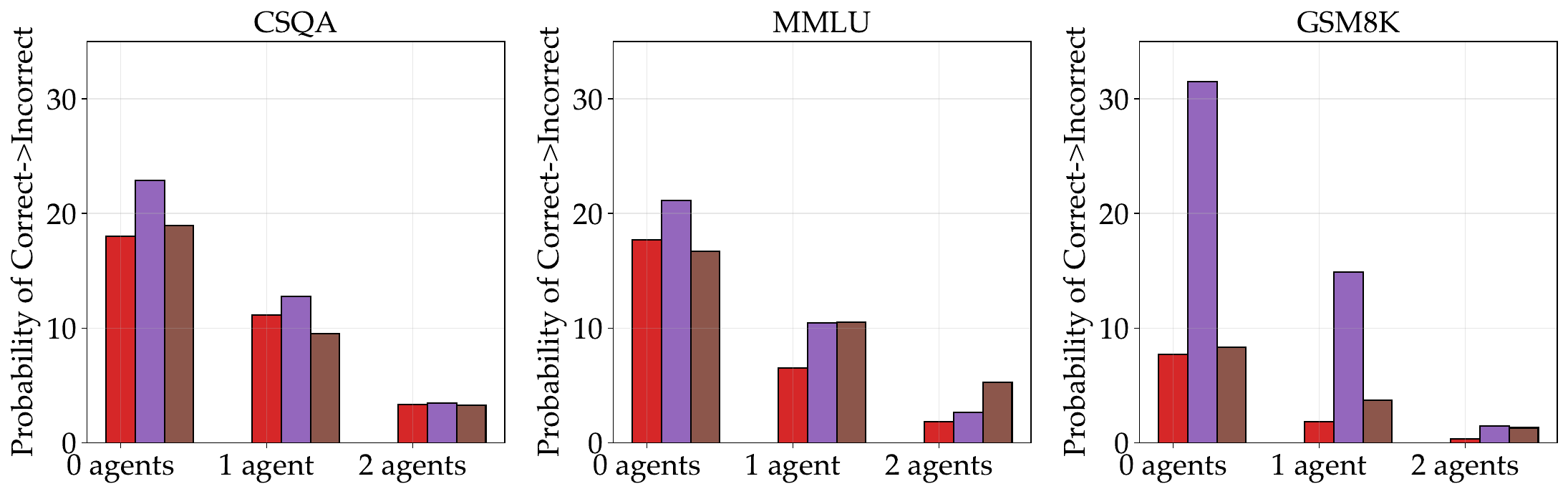}
    \caption{The likelihood of an answer being flipped from correct to incorrect (the \textit{undesirable} flip direction), plotted against the number of agents who agree with the ego agent, averaged across all rounds of debate. We find that the number of other agents who agree with the agent appears to be correlated with the frequency with which the agents flip their answers, indicating that the models may be influenced by social effects. We further observe significant variance in this answer-flipping behavior conditioned on the specific dataset or model in question. }
    \label{fig:agreement-ego-transitions-ci}
\end{figure}

We next proceed to investigate whether models, particularly stronger and more capable models, tend to be subject to social influence from disagreement with peers by examining the frequency of correct $\rightarrow$ incorrect answer flips. Figure \ref{fig:agreement-ego-transitions-ci} shows how frequently an agent changes its answer from correct $\rightarrow$ incorrect as a function of how many other agents initially agreed with it. Across all datasets, we observe that the likelihood of an undesirable answer-flip is highest when the ego agent is isolated -- i.e. when no other agents agree with its answer -- and decreases as more peers agree. Interestingly, this effect is highly variable across different datasets and models; for instance, on GSM8K, the strongest and weakest models (GPT and Mistral) are far less likely to make undesirable answer-flips than the third model (LLaMA). This provides strong preliminary evidence that LLM reasoning is influenced by social factors: models are sensitive to patterns of agreement and disagreement with peers, and their internal mechanisms for balancing correctness vs consensus may be quite fragile in the face of disagreement with peers. 

\subsection{Are agents sycophantic?}
\label{sec:investigation-debate-harm-factors}

\begin{figure}
    \centering
    \includegraphics[height=0.025\textheight]{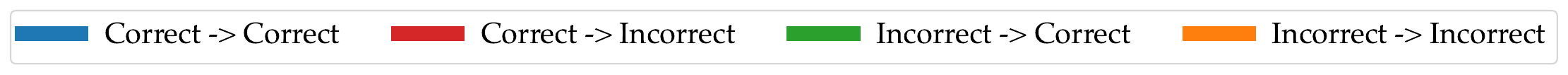}
    \includegraphics[height=0.025\textheight]{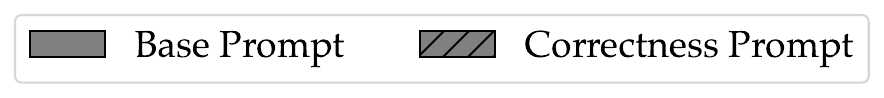}
    \includegraphics[width=0.8\textwidth]{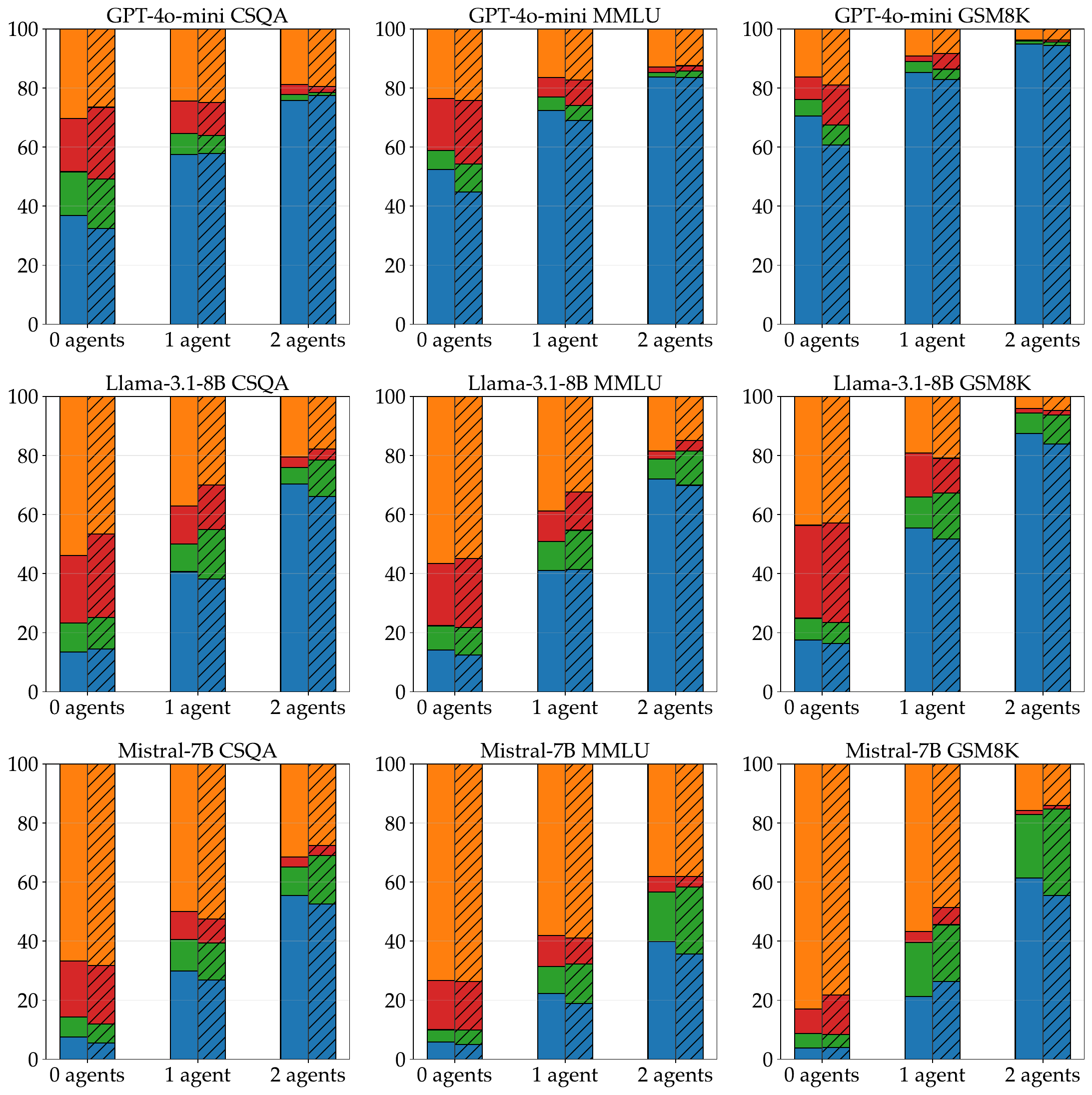}
    \caption{A comparison between the types of answer flips when agents are given the base prompt or the correctness payoff prompt. We find that introducing the correctness payoff intervention does not appear to decrease the number of undesirable \textcolor{red}{correct $\rightarrow$ incorrect flips} -- indicating that this intervention against sycophancy is insufficient on its own to resolve the issues with multi-agent debate.} 
    \label{fig:agreement-ego-transitions-correctness}
\end{figure}

We next investigate another potential contributing factor to failure within debate: models may exhibit \textit{sycophantic behavior} \citep{sharma2023understandingsycophancylanguagemodels} meaning the tendency, potentially due to RLHF \citep{rlhf}, of LLMs to prefer answers that match users' beliefs over correct answers. Specifically, we explore whether the degradation in debate performance could be due to a sycophancy bias expressed towards other LLM agents as opposed to users (that is, generalizing from the effects of RLHF.) We consider how agents respond to seeing the reasoning and answers from other agents over the course of debate by studying when models change their answer from correct $\rightarrow$ incorrect.  \footnote{\cite{yao2025peacemakertroublemakersycophancyshapes} also explore explore one particular form of sycophancy, showing that it does have some impact on the outcome of debate.}

\paragraph{Correctness payoff prompt.} Political scientists have found that partisan bias effects on question-answering accuracy \citep{bartels-partisan-bias} are significantly reduced when respondents are offered a monetary bonus for correct answers \citep{Bullock2015, prior2015}.  Inspired by these results, we design a new experiment intended to encourage LLM agents to prefer correctness over agreement with other agents, countering the sycophancy bias. In particular, we define a payoff function scoring the models based on the correctness of their answers and explicitly instruct them to maximize this payoff. We modify our prompts as follows:

\begin{tcolorbox}[
    colback=gray!5,
    colframe=gray!50,
    title={\textbf{System prompt}, $\mathcal{P}_{\text{system}}$},
    fonttitle=\bfseries,
    boxrule=1pt,
    arc=3pt,
    left=8pt,
    right=8pt,
    top=8pt,
    bottom=8pt
    ]
    You are a helpful assistant that can answer questions and complete tasks. You will receive a fixed payoff $p$ for each question given your response $X$ and the true answer $Y$,  $p = I[X==Y]$, where $I[condition]$ represents an indicator function which equals 1 if the condition is satisfied (i.e. your answer matches the true answer) and 0 otherwise. Your goal is to maximize your total payoff.
\end{tcolorbox}

\begin{tcolorbox}[
    colback=gray!5,
    colframe=gray!50,
    title={\textbf{System prompt}, $\mathcal{P}_{\text{debate}}$},
    fonttitle=\bfseries,
    boxrule=1pt,
    arc=3pt,
    left=8pt,
    right=8pt,
    top=8pt,
    bottom=8pt
    ]
    These are the solutions to the problem from other agents: \{AGENT\_RESPONSES\} Analyze your solution and that of other agents, provide an updated answer to maximize your payoff $p = I[X==Y]$, and explain your reasoning. Put your answer in the form (X) at the end of your response.
\end{tcolorbox}
We then investigate whether LLM agents are less likely to flip from correct to incorrect answers when they are prompted to maximize rewards for correct answers.

\textbf{Results.}  We present results in Fig. \ref{fig:agreement-ego-transitions-correctness}. We find that adding a payoff for correctness in the model prompt does not significantly reduce the likelihood that LLM agents flip their answers from correct to incorrect, regardless of the number of other agents that agree with them. In fact, we find that in many cases, the number of correct $\rightarrow$ incorrect transitions actually \textit{increases} when using the correctness-payoff prompt -- a counter-intuitive result that seems to suggest that asking models to prioritize correctness may not help resolve the issues observed in multi-agent debate. 

\subsection{Influence of Model and Task Type}

Across all our analyses (Figures \ref{fig:combined-sycophancy}, \ref{fig:agreement-ego-transitions-ci}, and \ref{fig:agreement-ego-transitions-correctness}), we observe substantial variation in behavior across individual models and tasks. Models of different individual abilities exhibit distinct patterns of answer change and sensitivity to peer disagreement, suggesting that an agent's capability itself can influence how social dynamics unfold. Likewise, the magnitude and direction of these effects differ across tasks such as CommonSenseQA, MMLU, and GSM8K, implying that task structure and domain complexity also shape how debates evolve. Together, these findings indicate that no single mechanism, such as sycophancy alone, can fully explain debate failures: rather, the interplay between model capability, task characteristics, and social influence jointly determines how and when agents revise their beliefs.

\section{Discussion}
\label{sec:discussion}

Our findings challenge the prevailing and natural view that deliberation among AI agents will always improve reasoning. In fact, our experiments reveal that multi-agent debate can sometimes degrade performance: group accuracy often declines over successive rounds of debate. This counterintuitive trend holds even when the majority of agents perform well individually on the task and even when weaker models have access to the reasoning of stronger models. In other words, additional exchange of reasons between agents does not always correct mistakes; instead, it may amplify them.

We study a number of correlated and contributing factors to these failure modes in debate, including sequential revision, social conditioning, and sycophancy. We additionally show across all experiments how models of different capabilities respond differently within multi-agent debate, and how the type of task or dataset can also have a significant influence on the results of debate. We find that beyond a simple single cause, multiple factors exist that likely contribute to failure modes in multi-agent debate, with the net effect that heterogeneous groups of models frequently converge on wrong answers together, potentially negating the benefits of multi-agent debate. 

These results suggest that naive debate protocols can inadvertently amplify errors rather than correcting them, underscoring the need for more principled approaches to multi-agent debate. Future debate frameworks should incorporate mechanisms that promote critical evaluation over consensus -- such as encouraging agents to assess the soundness of others' reasoning, integrating confidence estimates or credibility scores to weight contributions by expertise, and rewarding independent verification of claims. Training or incentive schemes might further discourage superficial agreement by penalizing unsupported conformity. Ultimately, improving the robustness of debate requires aligning models not merely to achieve consensus, but to engage in constructive epistemic disagreement -- challenging peers when warranted and maintaining justified beliefs under social pressure.

\section*{Acknowledgments and Disclosure of Funding}
Resources used in preparing this research were provided, in part, by the Province of Ontario, the Government of Canada through CIFAR, and companies sponsoring the Vector Institute \url{www.vectorinstitute.ai/#partners}.

\bibliography{example_paper}

\begin{thebibliography}{37}
\providecommand{\natexlab}[1]{#1}
\providecommand{\url}[1]{\texttt{#1}}
\expandafter\ifx\csname urlstyle\endcsname\relax
  \providecommand{\doi}[1]{doi: #1}\else
  \providecommand{\doi}{doi: \begingroup \urlstyle{rm}\Url}\fi

\bibitem[Agarwal and Khanna(2025)]{agarwal2025persuasionoverridestruthmultiagent}
Mahak Agarwal and Divyam Khanna.
\newblock When persuasion overrides truth in multi-agent llm debates: Introducing a confidence-weighted persuasion override rate (cw-por), 2025.
\newblock URL \url{https://arxiv.org/abs/2504.00374}.

\bibitem[Amayuelas et~al.(2024)Amayuelas, Yang, Antoniades, Hua, Pan, and Wang]{amayuelas2024multiagent}
Alfonso Amayuelas, Xianjun Yang, Antonis Antoniades, Wenyue Hua, Liangming Pan, and William Wang.
\newblock Multiagent collaboration attack: Investigating adversarial attacks in large language model collaborations via debate.
\newblock \emph{arXiv preprint arXiv:2406.14711}, 2024.

\bibitem[Bartels(2002)]{bartels-partisan-bias}
Larry~M. Bartels.
\newblock Beyond the running tally: Partisan bias in political perceptions.
\newblock \emph{Political Behavior}, 24\penalty0 (2):\penalty0 117--150, 2002.
\newblock ISSN 01909320, 15736687.
\newblock URL \url{http://www.jstor.org/stable/1558352}.

\bibitem[Brown et~al.(2020)Brown, Mann, Ryder, Subbiah, Kaplan, Dhariwal, Neelakantan, Shyam, Sastry, Askell, et~al.]{brown2020language}
Tom Brown, Benjamin Mann, Nick Ryder, Melanie Subbiah, Jared~D Kaplan, Prafulla Dhariwal, Arvind Neelakantan, Pranav Shyam, Girish Sastry, Amanda Askell, et~al.
\newblock Language models are few-shot learners.
\newblock \emph{Advances in neural information processing systems}, 33:\penalty0 1877--1901, 2020.

\bibitem[Bullock et~al.(2015)Bullock, Gerber, Hill, and Huber]{Bullock2015}
John~G. Bullock, Alan~S. Gerber, Seth~J. Hill, and Gregory~A. Huber.
\newblock Partisan bias in factual beliefs about politics.
\newblock \emph{Quarterly Journal of Political Science}, 10\penalty0 (4):\penalty0 519--578, 2015.
\newblock ISSN 1554-0626.
\newblock \doi{10.1561/100.00014074}.
\newblock URL \url{http://dx.doi.org/10.1561/100.00014074}.

\bibitem[Chan et~al.(2023)Chan, Chen, Su, Yu, Xue, Zhang, Fu, and Liu]{chan2023chateval}
Chi-Min Chan, Weize Chen, Yusheng Su, Jianxuan Yu, Wei Xue, Shanghang Zhang, Jie Fu, and Zhiyuan Liu.
\newblock Chateval: Towards better llm-based evaluators through multi-agent debate.
\newblock \emph{arXiv preprint arXiv:2308.07201}, 2023.

\bibitem[Cobbe et~al.(2021)Cobbe, Kosaraju, Bavarian, Chen, Jun, Kaiser, Plappert, Tworek, Hilton, Nakano, Hesse, and Schulman]{gsm8k}
Karl Cobbe, Vineet Kosaraju, Mohammad Bavarian, Mark Chen, Heewoo Jun, Lukasz Kaiser, Matthias Plappert, Jerry Tworek, Jacob Hilton, Reiichiro Nakano, Christopher Hesse, and John Schulman.
\newblock Training verifiers to solve math word problems, 2021.
\newblock URL \url{https://arxiv.org/abs/2110.14168}.

\bibitem[Du et~al.(2023)Du, Li, Torralba, Tenenbaum, and Mordatch]{du2023improvingfactualityreasoninglanguage}
Yilun Du, Shuang Li, Antonio Torralba, Joshua~B. Tenenbaum, and Igor Mordatch.
\newblock Improving factuality and reasoning in language models through multiagent debate, 2023.
\newblock URL \url{https://arxiv.org/abs/2305.14325}.

\bibitem[Estornell and Liu(2024)]{estornell2024multi}
Andrew Estornell and Yang Liu.
\newblock Multi-llm debate: Framework, principals, and interventions.
\newblock \emph{Advances in Neural Information Processing Systems}, 37:\penalty0 28938--28964, 2024.

\bibitem[Estornell et~al.(2025)Estornell, Ton, Yao, and Liu]{estornell2025acccollabactorcriticapproachmultiagent}
Andrew Estornell, Jean-Francois Ton, Yuanshun Yao, and Yang Liu.
\newblock Acc-collab: An actor-critic approach to multi-agent llm collaboration, 2025.
\newblock URL \url{https://arxiv.org/abs/2411.00053}.

\bibitem[et~al.(2024)]{llama3}
Aaron~Grattafiori et~al.
\newblock The llama 3 herd of models, 2024.
\newblock URL \url{https://arxiv.org/abs/2407.21783}.

\bibitem[Fu et~al.(2023)Fu, Peng, Khot, and Lapata]{fu2023improving}
Yao Fu, Hao Peng, Tushar Khot, and Mirella Lapata.
\newblock Improving language model negotiation with self-play and in-context learning from ai feedback.
\newblock \emph{arXiv preprint arXiv:2305.10142}, 2023.

\bibitem[He et~al.(2025)He, Zou, Li, Chen, Xing, and Ma]{he2025enhancingllmreasoningmultipath}
Chengbo He, Bochao Zou, Xin Li, Jiansheng Chen, Junliang Xing, and Huimin Ma.
\newblock Enhancing llm reasoning with multi-path collaborative reactive and reflection agents, 2025.
\newblock URL \url{https://arxiv.org/abs/2501.00430}.

\bibitem[Hendrycks et~al.(2021)Hendrycks, Burns, Basart, Zou, Mazeika, Song, and Steinhardt]{mmlu}
Dan Hendrycks, Collin Burns, Steven Basart, Andy Zou, Mantas Mazeika, Dawn Song, and Jacob Steinhardt.
\newblock Measuring massive multitask language understanding, 2021.
\newblock URL \url{https://arxiv.org/abs/2009.03300}.

\bibitem[Huang et~al.(2023)Huang, Chen, Mishra, Zheng, Yu, Song, and Zhou]{huang2023large}
Jie Huang, Xinyun Chen, Swaroop Mishra, Huaixiu~Steven Zheng, Adams~Wei Yu, Xinying Song, and Denny Zhou.
\newblock Large language models cannot self-correct reasoning yet.
\newblock \emph{arXiv preprint arXiv:2310.01798}, 2023.

\bibitem[Irving et~al.(2018)Irving, Christiano, and Amodei]{irving2018aisafetydebate}
Geoffrey Irving, Paul Christiano, and Dario Amodei.
\newblock Ai safety via debate, 2018.
\newblock URL \url{https://arxiv.org/abs/1805.00899}.

\bibitem[Jiang et~al.(2023)Jiang, Sablayrolles, Mensch, Bamford, Chaplot, de~las Casas, Bressand, Lengyel, Lample, Saulnier, Lavaud, Lachaux, Stock, Scao, Lavril, Wang, Lacroix, and Sayed]{mistral}
Albert~Q. Jiang, Alexandre Sablayrolles, Arthur Mensch, Chris Bamford, Devendra~Singh Chaplot, Diego de~las Casas, Florian Bressand, Gianna Lengyel, Guillaume Lample, Lucile Saulnier, Lélio~Renard Lavaud, Marie-Anne Lachaux, Pierre Stock, Teven~Le Scao, Thibaut Lavril, Thomas Wang, Timothée Lacroix, and William~El Sayed.
\newblock Mistral 7b, 2023.
\newblock URL \url{https://arxiv.org/abs/2310.06825}.

\bibitem[Kamoi et~al.(2024)Kamoi, Zhang, Zhang, Han, and Zhang]{kamoi2024can}
Ryo Kamoi, Yusen Zhang, Nan Zhang, Jiawei Han, and Rui Zhang.
\newblock When can llms actually correct their own mistakes? a critical survey of self-correction of llms.
\newblock \emph{Transactions of the Association for Computational Linguistics}, 12:\penalty0 1417--1440, 2024.

\bibitem[Kaufmann et~al.(2024)Kaufmann, Weng, Bengs, and Hüllermeier]{rlhf}
Timo Kaufmann, Paul Weng, Viktor Bengs, and Eyke Hüllermeier.
\newblock A survey of reinforcement learning from human feedback, 2024.
\newblock URL \url{https://arxiv.org/abs/2312.14925}.

\bibitem[Kenton et~al.(2024)Kenton, Siegel, Kramár, Brown-Cohen, Albanie, Bulian, Agarwal, Lindner, Tang, Goodman, and Shah]{kenton2024scalableoversightweakllms}
Zachary Kenton, Noah~Y. Siegel, János Kramár, Jonah Brown-Cohen, Samuel Albanie, Jannis Bulian, Rishabh Agarwal, David Lindner, Yunhao Tang, Noah~D. Goodman, and Rohin Shah.
\newblock On scalable oversight with weak llms judging strong llms, 2024.
\newblock URL \url{https://arxiv.org/abs/2407.04622}.

\bibitem[Khan et~al.(2024)Khan, Hughes, Valentine, Ruis, Sachan, Radhakrishnan, Grefenstette, Bowman, Rocktäschel, and Perez]{khan2024debatingpersuasivellmsleads}
Akbir Khan, John Hughes, Dan Valentine, Laura Ruis, Kshitij Sachan, Ansh Radhakrishnan, Edward Grefenstette, Samuel~R. Bowman, Tim Rocktäschel, and Ethan Perez.
\newblock Debating with more persuasive llms leads to more truthful answers, 2024.
\newblock URL \url{https://arxiv.org/abs/2402.06782}.

\bibitem[Li et~al.(2023)Li, Hammoud, Itani, Khizbullin, and Ghanem]{li2023camelcommunicativeagentsmind}
Guohao Li, Hasan Abed Al~Kader Hammoud, Hani Itani, Dmitrii Khizbullin, and Bernard Ghanem.
\newblock Camel: Communicative agents for "mind" exploration of large language model society, 2023.
\newblock URL \url{https://arxiv.org/abs/2303.17760}.

\bibitem[Liang et~al.(2023)Liang, He, Jiao, Wang, Wang, Wang, Yang, Shi, and Tu]{liang2023encouraging}
Tian Liang, Zhiwei He, Wenxiang Jiao, Xing Wang, Yan Wang, Rui Wang, Yujiu Yang, Shuming Shi, and Zhaopeng Tu.
\newblock Encouraging divergent thinking in large language models through multi-agent debate.
\newblock \emph{arXiv preprint arXiv:2305.19118}, 2023.

\bibitem[Madaan et~al.(2023)Madaan, Tandon, Gupta, Hallinan, Gao, Wiegreffe, Alon, Dziri, Prabhumoye, Yang, et~al.]{madaan2023self}
Aman Madaan, Niket Tandon, Prakhar Gupta, Skyler Hallinan, Luyu Gao, Sarah Wiegreffe, Uri Alon, Nouha Dziri, Shrimai Prabhumoye, Yiming Yang, et~al.
\newblock Self-refine: Iterative refinement with self-feedback.
\newblock \emph{Advances in Neural Information Processing Systems}, 36:\penalty0 46534--46594, 2023.

\bibitem[Michael et~al.(2023)Michael, Mahdi, Rein, Petty, Dirani, Padmakumar, and Bowman]{michael2023debatehelpssuperviseunreliable}
Julian Michael, Salsabila Mahdi, David Rein, Jackson Petty, Julien Dirani, Vishakh Padmakumar, and Samuel~R. Bowman.
\newblock Debate helps supervise unreliable experts, 2023.
\newblock URL \url{https://arxiv.org/abs/2311.08702}.

\bibitem[OpenAI(2024)]{openai2024gpt4omini}
OpenAI.
\newblock Gpt-4o mini: Advancing cost-efficient intelligence.
\newblock \url{https://openai.com/index/gpt-4o-mini-advancing-cost-efficient-intelligence/}, 2024.

\bibitem[Prior et~al.(2015)Prior, Sood, and Khanna]{prior2015}
Markus Prior, Gaurav Sood, and Kabir Khanna.
\newblock You cannot be serious: The impact of accuracy incentives on partisan bias in reports of economic perceptions.
\newblock \emph{Quarterly Journal of Political Science}, 10\penalty0 (4):\penalty0 489--518, December 2015.
\newblock \doi{10.1561/100.00014127}.
\newblock URL \url{https://ideas.repec.org/a/now/jlqjps/100.00014127.html}.

\bibitem[Sharma et~al.(2023)Sharma, Tong, Korbak, Duvenaud, Askell, Bowman, Cheng, Durmus, Hatfield-Dodds, Johnston, Kravec, Maxwell, McCandlish, Ndousse, Rausch, Schiefer, Yan, Zhang, and Perez]{sharma2023understandingsycophancylanguagemodels}
Mrinank Sharma, Meg Tong, Tomasz Korbak, David Duvenaud, Amanda Askell, Samuel~R. Bowman, Newton Cheng, Esin Durmus, Zac Hatfield-Dodds, Scott~R. Johnston, Shauna Kravec, Timothy Maxwell, Sam McCandlish, Kamal Ndousse, Oliver Rausch, Nicholas Schiefer, Da~Yan, Miranda Zhang, and Ethan Perez.
\newblock Towards understanding sycophancy in language models, 2023.
\newblock URL \url{https://arxiv.org/abs/2310.13548}.

\bibitem[Subramaniam et~al.(2025)Subramaniam, Du, Tenenbaum, Torralba, Li, and Mordatch]{subramaniam2025multiagent}
Vighnesh Subramaniam, Yilun Du, Joshua~B Tenenbaum, Antonio Torralba, Shuang Li, and Igor Mordatch.
\newblock Multiagent finetuning: Self improvement with diverse reasoning chains.
\newblock \emph{arXiv preprint arXiv:2501.05707}, 2025.

\bibitem[Talmor et~al.(2019)Talmor, Herzig, Lourie, and Berant]{commonsenseqa}
Alon Talmor, Jonathan Herzig, Nicholas Lourie, and Jonathan Berant.
\newblock Commonsenseqa: A question answering challenge targeting commonsense knowledge, 2019.
\newblock URL \url{https://arxiv.org/abs/1811.00937}.

\bibitem[Tran et~al.(2025)Tran, Dao, Nguyen, Pham, O'Sullivan, and Nguyen]{tran2025multi}
Khanh-Tung Tran, Dung Dao, Minh-Duong Nguyen, Quoc-Viet Pham, Barry O'Sullivan, and Hoang~D Nguyen.
\newblock Multi-agent collaboration mechanisms: A survey of llms.
\newblock \emph{arXiv preprint arXiv:2501.06322}, 2025.

\bibitem[Wang et~al.(2024)Wang, Wang, Athiwaratkun, Zhang, and Zou]{wang2024mixture}
Junlin Wang, Jue Wang, Ben Athiwaratkun, Ce~Zhang, and James Zou.
\newblock Mixture-of-agents enhances large language model capabilities.
\newblock \emph{arXiv preprint arXiv:2406.04692}, 2024.

\bibitem[Wang et~al.(2022)Wang, Wei, Schuurmans, Le, Chi, Narang, Chowdhery, and Zhou]{wang2022self}
Xuezhi Wang, Jason Wei, Dale Schuurmans, Quoc Le, Ed~Chi, Sharan Narang, Aakanksha Chowdhery, and Denny Zhou.
\newblock Self-consistency improves chain of thought reasoning in language models.
\newblock \emph{arXiv preprint arXiv:2203.11171}, 2022.

\bibitem[Wang et~al.(2023)Wang, Wei, Schuurmans, Le, Chi, Narang, Chowdhery, and Zhou]{wang2023selfconsistencyimproveschainthought}
Xuezhi Wang, Jason Wei, Dale Schuurmans, Quoc Le, Ed~Chi, Sharan Narang, Aakanksha Chowdhery, and Denny Zhou.
\newblock Self-consistency improves chain of thought reasoning in language models, 2023.
\newblock URL \url{https://arxiv.org/abs/2203.11171}.

\bibitem[Wu et~al.(2023)Wu, Bansal, Zhang, Wu, Li, Zhu, Jiang, Zhang, Zhang, Liu, Awadallah, White, Burger, and Wang]{wu2023autogenenablingnextgenllm}
Qingyun Wu, Gagan Bansal, Jieyu Zhang, Yiran Wu, Beibin Li, Erkang Zhu, Li~Jiang, Xiaoyun Zhang, Shaokun Zhang, Jiale Liu, Ahmed~Hassan Awadallah, Ryen~W White, Doug Burger, and Chi Wang.
\newblock Autogen: Enabling next-gen llm applications via multi-agent conversation, 2023.
\newblock URL \url{https://arxiv.org/abs/2308.08155}.

\bibitem[Yang et~al.(2025)Yang, Chen, Siew, Lorido-Botran, and Joe-Wong]{yang2025llmpowereddecentralizedgenerativeagents}
Hanqing Yang, Jingdi Chen, Marie Siew, Tania Lorido-Botran, and Carlee Joe-Wong.
\newblock Llm-powered decentralized generative agents with adaptive hierarchical knowledge graph for cooperative planning, 2025.
\newblock URL \url{https://arxiv.org/abs/2502.05453}.

\bibitem[Yao et~al.(2025)Yao, Shang, Du, He, Lian, Zhang, Su, Swamy, and Qi]{yao2025peacemakertroublemakersycophancyshapes}
Binwei Yao, Chao Shang, Wanyu Du, Jianfeng He, Ruixue Lian, Yi~Zhang, Hang Su, Sandesh Swamy, and Yanjun Qi.
\newblock Peacemaker or troublemaker: How sycophancy shapes multi-agent debate, 2025.
\newblock URL \url{https://arxiv.org/abs/2509.23055}.

\end{thebibliography}
\bibliographystyle{plainnat}

\appendix

\end{document}